\documentclass[letterpaper, 10pt, conference]{ieeeconf} 

\IEEEoverridecommandlockouts                    
                                                          
\usepackage{multirow}

\usepackage{algpseudocode}

\usepackage{rotating}
\usepackage{color}
\usepackage{subfigure}
\usepackage{algorithm}
\usepackage[justification=centering]{caption}
\usepackage{braket}
\usepackage{amssymb}

\definecolor{darkgreen}{rgb}{0,0.8,0}

\title{CINet: A Learning Based Approach to Incremental Context Modeling in Robots}

\author{Fethiye Irmak Do\u{g}an$^{1,\star}$, \.{I}lker Bozcan$^{1,\star}$, Mehmet \c{C}elik$^1$ and Sinan Kalkan$^{1}$
\\ *$^\star$ Equal contribution
\thanks{$^{1}$All authors are with the KOVAN Research Lab at the Department of Computer Engineering, Middle East Technical University, Ankara, Turkey
        {\tt\small \{irmak.dogan, ilker.bozcan, mcelik, skalkan\}@metu.edu.tr}}%
}

\begin{document}

\maketitle
\thispagestyle{empty}
\pagestyle{empty}

\begin{abstract}
There have been several attempts at modeling context in robots. However, either these attempts assume a fixed number of contexts or use a rule-based approach to determine when to increment the number of contexts. In this paper, we pose the task of when to increment as a learning problem, which we solve using a Recurrent Neural Network. We show that the network successfully (with 98\% testing accuracy) learns to predict when to increment, and demonstrate, in a scene modeling problem (where the correct number of contexts is not known), that the robot increments the number of contexts in an expected manner (i.e., the entropy of the system is reduced). We also present how the incremental model can be used for various scene reasoning tasks.
\end{abstract}

\section{Introduction}

Context is known to be very crucial for human cognition, functioning as a modulator affecting our perception, reasoning, communication and action \cite{yeh2006situated,barsalou2009simulation}. The robots that we expect to have an important role in our daily lives in the near future should have the ability to perceive, to learn and to use context, like we do. 

However, \textit{context modeling should be incremental} since it is not possible know beforehand the set of all possible situations that a robot is going to encounter. The types of situations (contexts) are going to be different even for a simple vacuum cleaning robot based on the environment and the users.

Although there have been incremental context modeling efforts on robots \cite{ortiz2014incremental,CelikkanatContext2014} or incremental topic modeling in linguistics \cite{canini2009online}, to the best of our knowledge, this is the first study that considers the question of ``when to increment the number of contexts'' as a \textbf{learning problem}. However, this is challenging since there is no ground truth on the correct number of contexts in any problem domain.

\begin{figure}

\centerline{
	\includegraphics[width=\columnwidth]{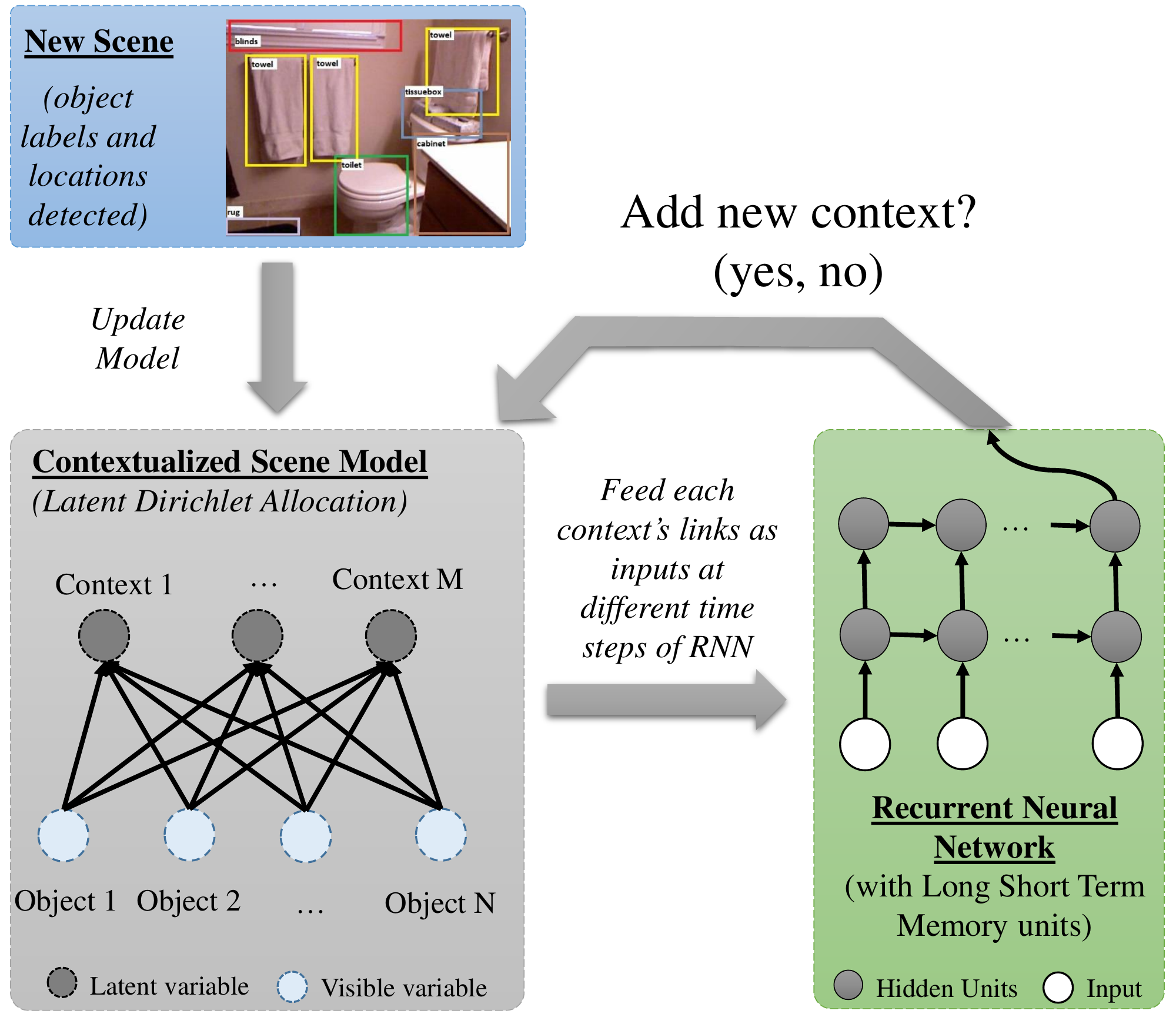}}
    \caption{An overview of how we address incremental context modeling as a learning problem. When a new scene is encountered, the objects are detected (not a contribution of the paper), and the Latent Dirichlet Allocation Model is updated. The updated model is fed as input to a Recurrent Neural Network, which predicts whether to increment the number of contexts or not. \label{fig:overview}}
\end{figure}

\subsection{Related Work}

\textbf{Scene modeling:} Scene modeling is the task of modeling what is in the scene. Such scene models are critical in robots since they allow reasoning about the scene and the objects in it. Many models have been proposed in computer vision and robotics using probabilistic models such as Markov Random Fields \cite{anand2013contextually,CelikkanatConceptWeb2014}, Bayesian Networks \cite{sheikh2005bayesian,li2017context}, Latent Dirichlet Allocation variants \cite{wang2008spatial, Philbin08a}, predicate logic  \cite{mastrogiovanni2011robots,hwang2006ontology}, and Scene Graphs \cite{blumenthal2014towards}. There have also been many attempts for ontology-based scene modeling where objects and various types of relations are modeled \cite{hwang2006ontology,saxena2014robobrain,tenorth2009knowrob}.

\textbf{Context modeling:}
Although context has been widely studied in other fields, it has not received sufficient attention in the robotics community, except for, e.g., \cite{anand2013contextually}
who used spatial relations between objects as contextual prior for object detection in a Markov Random Field; \cite{CelikkanatContext2014}, who adapted Latent Dirichlet Allocation on top of object concepts for rule-based incremental context modeling; and \cite{li2017context}, who proposed using a variant of Bayesian Networks for context modeling in underwater robots.

\textbf{Incremental context or topic modeling:}
There have been some attempts at incremental context or topic modeling, in text modeling \cite{canini2009online}, computer vision \cite{yu2015incremental} and in robotics \cite{ortiz2014incremental,CelikkanatContext2014}. However, these attempts are generally rule-based approaches, looking at the errors or the entropy (perplexity) of the system to determine when to increment. Since these rules are hand-crafted by intuition, they may fail to capture the underlying structure of the data for determining when to increment. There are also methods such as Hierarchical Dirichlet Processes \cite{TehEtAl2006} or its nested version \cite{paisley2015nested} that assume the availability of all data to estimate the number of topics or assume an infinite number of topics, which are both unrealistic for robots continually interacting with the environment and getting into new contexts through their lifetime.

\textbf{In summary}, we notice that there are no studies that address learnability of incrementing the number of topics, which is a necessity for life-long learning robots.

\subsection{Contributions}

The main contributions of our work are the following: (i) \textbf{Pose incremental context modeling as a learning problem}. This is challenging since it requires a training dataset with the correct number of contexts to train a model and there is no such dataset available. We solve this issue by using Latent Dirichlet Allocation \cite{griffiths2004finding}, which, being a generative model, allows one to generate artificial data with a given number of contexts. For any data generated with a certain number of contexts, we can now know the correct number of contexts and train a model. (ii) \textbf{Solve this learning problem by using a deep network}. For this end, we employ a Recurrent Neural Network, and model the problem as a sequence-to-label problem. The input of the network at each time step is the weights of a context to the objects, and the output is a binary decision on whether to increment the number of topics or not -- See also Figure \ref{fig:overview}.

On an artificial dataset and a real dataset (SUN-RGBD scene dataset \cite{song2015sun}, we show that the network learns to add a new context when it decides necessary on encountering new scenes. Moreover, we compared our method with another incremental Latent Dirichlet Allocation method \cite{CelikkanatContext2014} and incremental Boltzmann Machines \cite{DoganEtAl2018}, and demonstrated that it performs better.

\section{Learning-based Incremental Context Modeling}

We assume that objects in scenes occur in different contexts, and we can model such contexts as latent variables defined over the objects as shown in Figure \ref{fig:overview} and in our previous work \cite{CelikkanatContext2014}.

\subsection{Contextualized Scene Modeling with Latent Dirichlet Allocation (LDA)}

LDA \cite{griffiths2004finding} is a generative model widely used for topic modeling in document classification. Since it is more intuitive, we will introduce LDA from a document modeling perspective: Assuming a document $d \in \mathbb{D}$ is a set of words $w_1, ..., w_N$ drawn from a fixed vocabulary ($w_i \in \mathbb{W}$ for vocabulary of size $|\mathbb{W}|$), LDA posits a finite mixture over a fixed set of topics $z_1, ..., z_k$, ($z_t \in \mathbb{Z}$ with $|\mathbb{Z}|=k$ being the topic count). Then, a document can be described by the probability of relating to these topics, $p(z_t | d_i)$. Conversely, a topic is modeled by the likelihood for a document of this topic to contain each word in the vocabulary, $p(w_j  | z_t)$. LDA proposes to infer these document and topic probability distributions from a set of documents $\mathbb{D}$.

For contextualized scene modeling, we replace a document $d_i$ with a scene $s_i$, a word $w_i$ with an object $o_i$, and a topic $t_i$ with a context $c_i$, as suggested by our previous work \cite{CelikkanatContext2014}.

\subsection{Dataset Collection}
\label{sect:dataset}
There is no dataset with correct number of contexts, and it is not sensible to take the number of high-level categories in an existing dataset (such as those used in scene classification) as contexts since there may be more contexts in the data than the number of high-level categories. 

\begin{figure}[hb!]
\centerline{
		\includegraphics[width=0.45\textwidth]{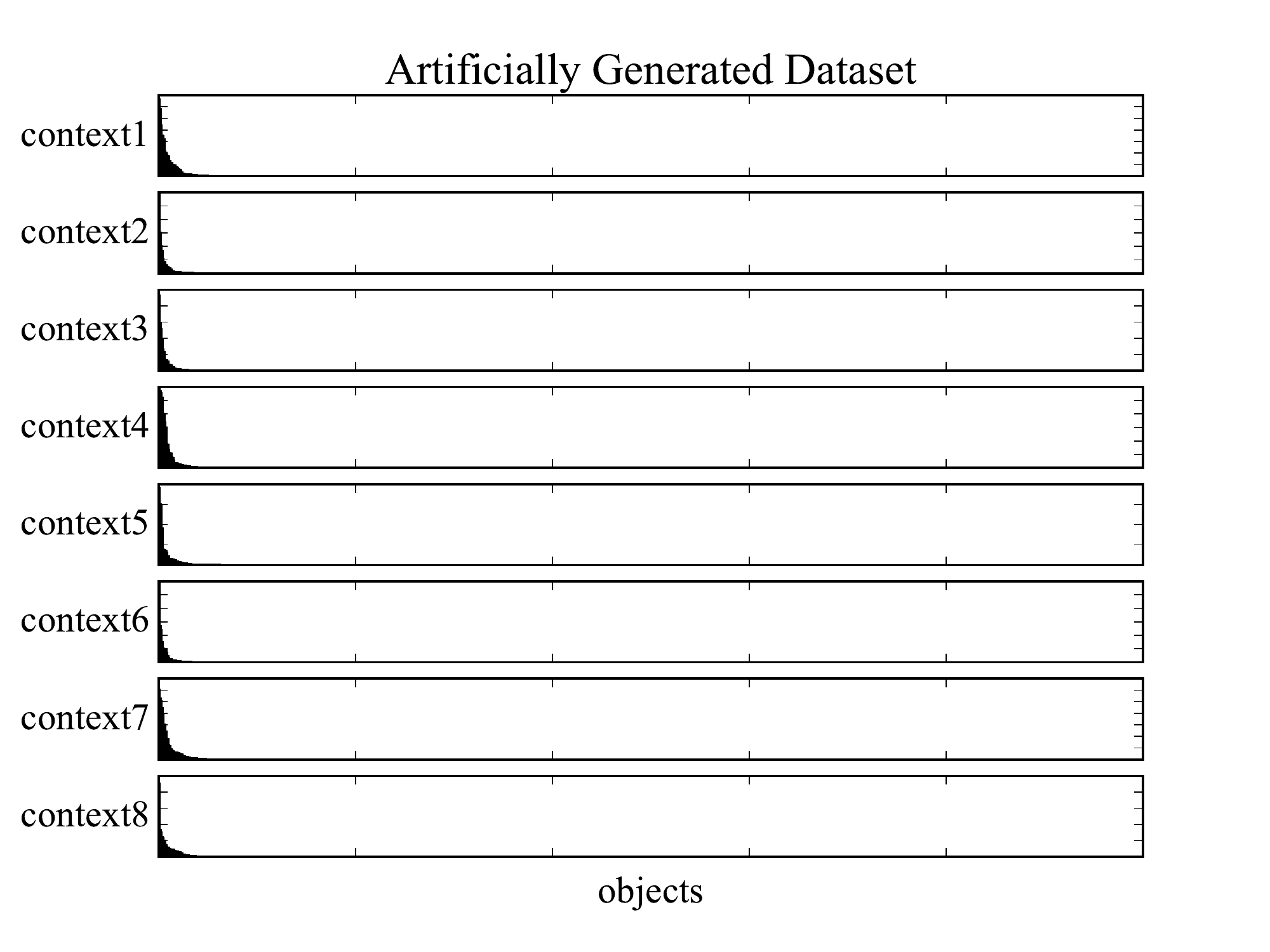}
}
\centerline{
		\includegraphics[width=0.45\textwidth]{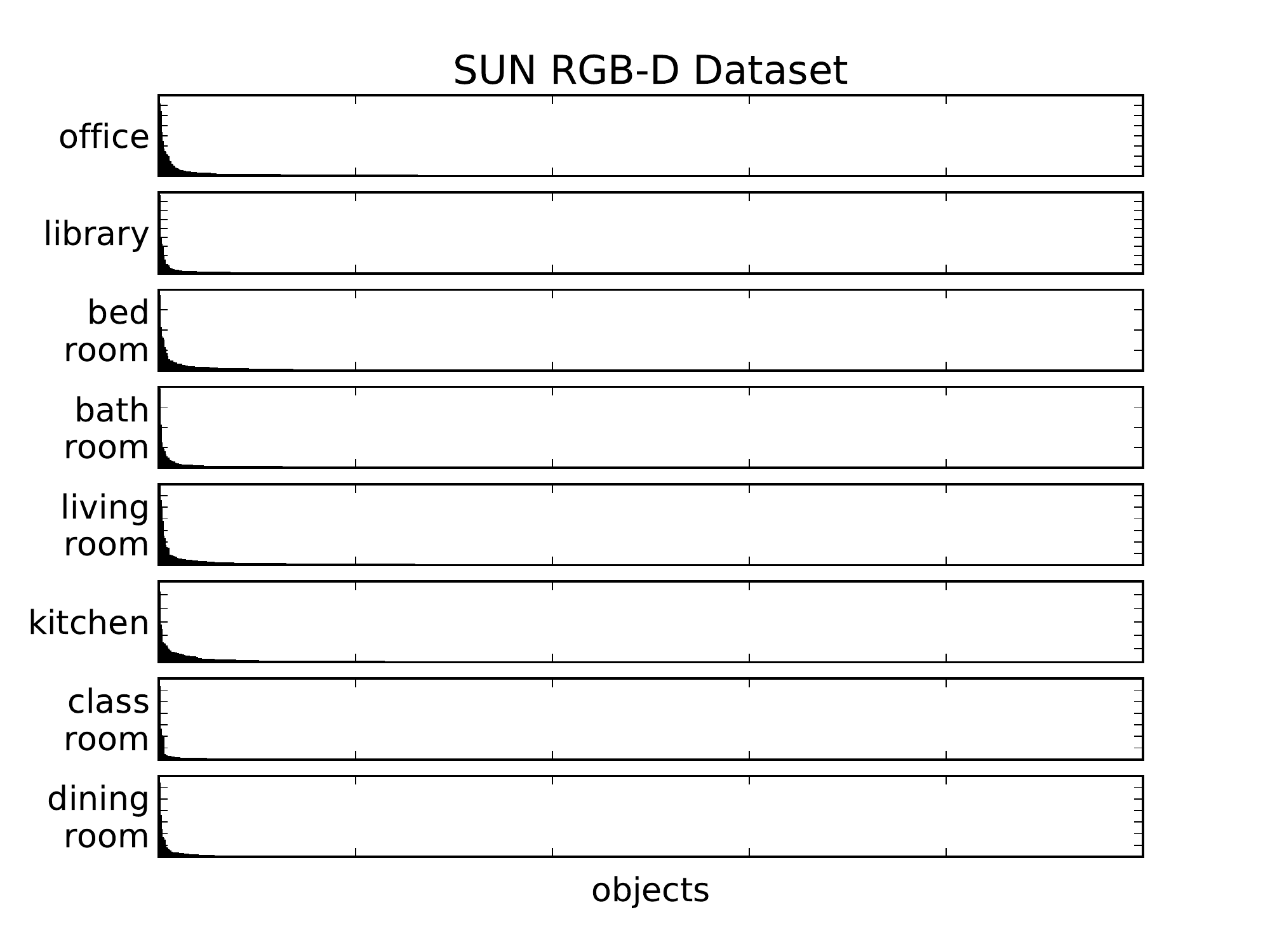}
}
\caption{Context-object co-occurrence frequencies in the artificially generated dataset (top) and the real dataset, i.e., the SUN-RGBD dataset \cite{song2015sun} (bottom). We see that the distribution of the artificial dataset and that of the real dataset are similar, suggesting that a model trained on the artificial dataset may generalize to the real dataset. \label{fig:distributions}}
\end{figure}

\begin{figure*}[hbt!]
   \vspace*{1.0cm}
\centerline{
	\includegraphics[width=0.649\textwidth]{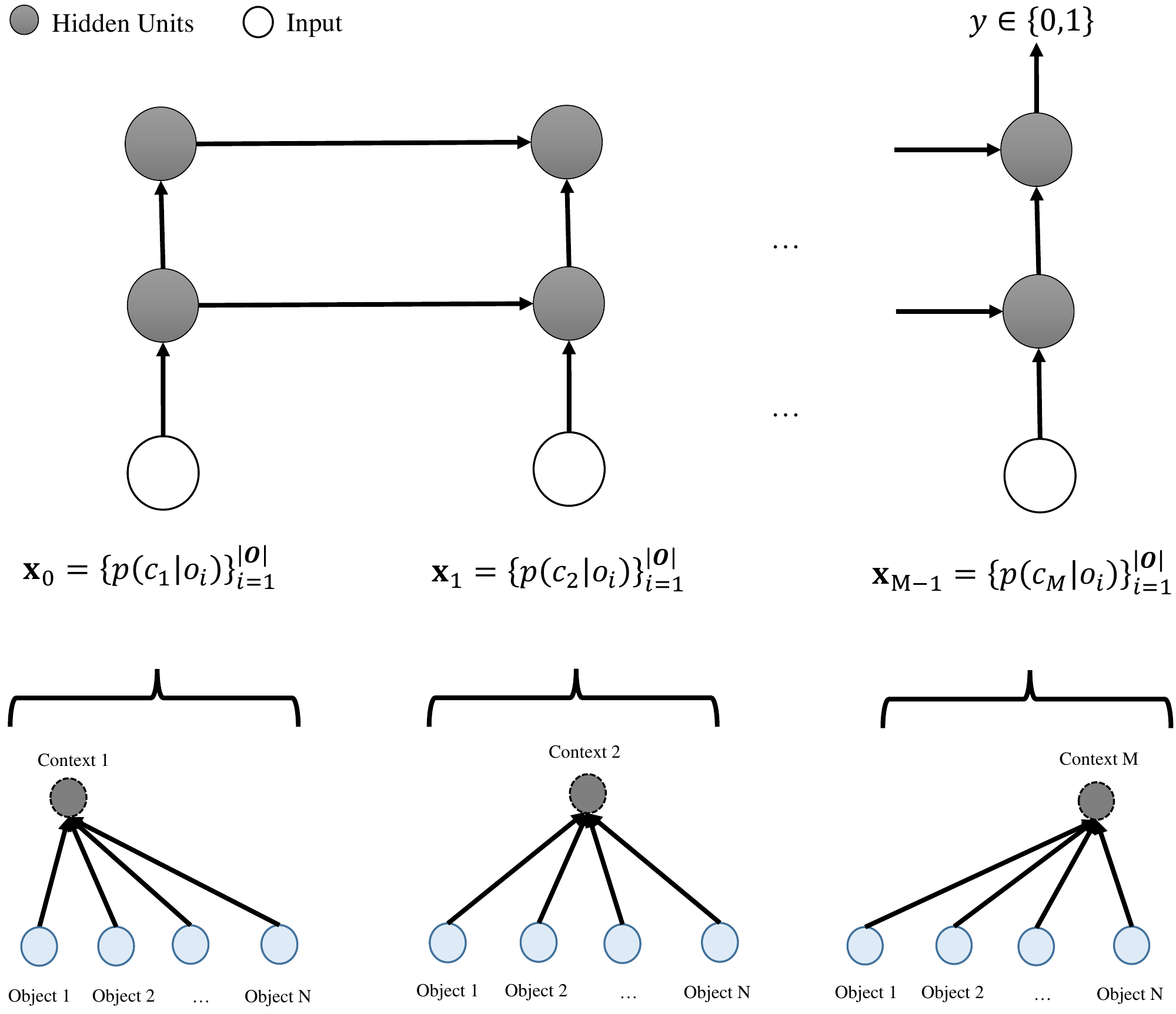}
}
\caption{An overall view of the unfolded RNN architecture used in the paper. Various types and number of hidden units, and number of layers have been experimented. \label{fig:architecture}}
\end{figure*}

To address this issue, we used Latent Dirichlet Allocation (LDA) \cite{griffiths2004finding}, which, being a generative model, allows one to sample artificial data with various number of contexts (topics). Since we are interested in the distribution between the contexts and the objects, and whether we can determine when to increment the number of topics for an online model, the data being artificial does not pose a problem -- as shown by our experiments. The only assumption we make here is that the contexts follow a Dirichlet distribution, which is a reasonable approximation since it is a family of distributions that can approximate the various types of distributions many high-level categories and natural phenomena (e.g., objects, words, scenes, activities) may follow; e.g., power-law \cite{markovic2014power}. Looking at the distribution of objects in contexts generated by the model and the objects in scenes in a real dataset, we see that the distributions align closely as shown in Figure \ref{fig:distributions}.

In LDA, a Dirichlet prior $\alpha$, describing the corpus (environment), is assumed. A parameter, $\theta$, governing the distribution over the contexts is sampled from Dir($\alpha$). The number of objects, $N$, in a scene is randomly sampled from a suitable distribution, e.g., Poisson. Then, for each scene $S$ to be generated, a topic (context) $c_i$ is sampled from Discrete($\theta$) for each index $i$, and an object $o_i$ is sampled using $p(o_i | c_i, \beta)$. We selected $\alpha$ as 0.9, and $\beta$ as 0.01, since they yielded matching distributions for the real data.

Let us use $D^k=\{S^k_1, ..., S^k_{L_k}\}$ to denote the dataset of scenes generated with $k$ number of contexts. We can now train LDA models with $k_0$ contexts s.t. $k_0\le k$. The training dataset of tuples ($\mathbf{x},y$) used for training the deep network can be constructed as follows:
\begin{itemize}
\item $\mathbf{x}$: The input vector to the network, describing the current state of the LDA model. Since the number of contexts (hidden topics) is not fixed beforehand, $\mathbf{x}$ is variable length. Therefore, we take $\mathbf{x}$ to be a sequence of sub-vectors $\mathbf{x}_i = \mathbf{p}_{c_i}=\{p(c_i | o_j)\}_{j=1}^{O}$, i.e., a sequence of conditional probabilities of each context given an object. If an object is not used at all during the generation of any scene, the probability for that object is set to zero. We also considered using $\mathbf{p}_{o_i}=\{p(o_i | c_j)\}_{j=1}^{C}$, i.e., a vector of probabilities of each object given a context.
\item $y$: The expected output of the network; a binary variable (0 or 1) describing whether to increment ($y=1$) the number of contexts or not ($y=0$).
\end{itemize}

In total, we collected 14400 scenes (documents) from 1000 objects (vocabulary) with 10 different contexts (from 1 to 10). Each scene had 100 objects. From various combinations of these instances, we gathered 27000 ($\mathbf{x},y$) pairs for training and 3400 for testing. The limitation for gathering more data was the long computation time; for each instance, LDA model needed to be trained. 

\begin{table*}
\centering
\vspace*{1.0cm}
\caption{Training and test accuracies for the different models of CINet. Accuracy is the percentage of correct increment decisions calculated over the artificial data. \label{tbl:accuracies}}
\begin{tabular}{|l|cc|cc|cc|} \hline
   & \multicolumn{2}{|c|}{\textbf{Input:} $\mathbf{p}_{c_i}$\vspace*{0.1cm}} & \multicolumn{2}{|c|}{\textbf{Input:} $\mathbf{p}_{o_i}$} & \multicolumn{2}{|c|}{\textbf{Input:} $\mathbf{p}_{o_i}\oplus \mathbf{p}_{c_i}$}  \\
   &  \textbf{Training Acc.} & \textbf{Test Acc.} &  \textbf{Training Acc.} & \textbf{Test Acc.}    &  \textbf{Training Acc.} & \textbf{Test Acc.} \\ \hline\hline
Vanilla RNN (1 layers)  & 98.0\%    & 97.1\% 	& 72.6\% & 66.2\%		& 99.3\% & 95.5\%	\\
Vanilla RNN (2 layers)  & 99.2\%    & 97.7\%  	& 97.0\% & 69.5\%		& 97.5\% & 94.8\%	\\
Vanilla RNN (3 layers)  & 99.7\%    & 97.9\%  	& 99.5\% & 71.4\%		& 94.8\% & 93.9\%	\\ \hline
GRU  (1 layers) 		& 99.4\%    & 94.7\%  	& 97.2\% & 71.0\%		& 99.5\% & 93.7\%	\\
GRU  (2 layers) 		& 99.3\%    & 97.3\%  	& 99.9\% & 71.7\%		& 99.4\% & 96.0\%	\\
GRU  (3 layers) 		& 99.4\%    & 97.3\%  	& 99.8\% & 71.7\%		& 94.3\% & 94.2\%	\\ \hline
LSTM (1 layers) 		& 99.1\%    & 92.9\%  	& 73.4\% & 67.5\%		& 99.7\% & 89.6\%	\\
LSTM (2 layers) 		& 99.4\%    & 97.9\%  	& 99.7\% & 70.6\%		& 99.7\% & 96.7\%	\\
LSTM (3 layers) & \textbf{99.9}\%  & \textbf{98.0\%}  	& 99.4\% & 70.9\%		& 99.4\% & 94.2\%	\\ \hline
\end{tabular}
\end{table*}

\subsection{Context Incrementing Network (CINet)}

Since our input vectors $\mathbf{x}$ have varying lengths, we used Recurrent Neural Networks (RNNs) to cast the problem as a learning problem, where, given $\mathbf{x}$ describing the state of the LDA model, $y$ (whether to increment the number of contexts) is predicted. An overall view of a CINet architecture is provided in Figure \ref{fig:architecture}. We evaluated different types (LSTM \cite{hochreiter1997long}, GRU \cite{cho2014properties}) and number of hidden units and layers, and reported only the most competing ones in the next section.

For estimating the error, we formulated a binary cross-entropy loss $\mathcal{J}$ which CINet is expected to minimize:
\begin{equation}
\mathcal{J}(W)=-\frac{1}{n}\sum_i \left[ y_{i}\log \hat{y}_{i} + (1-y_i)\log (1-\hat{y}_i)\right],
\end{equation}
{\noindent}where $W$ is the set of parameters in the network; $\hat{y}_i$ is the prediction of the network for the $i^{th}$ sample; and $n$ is the number of samples in the dataset or the batch. As a precaution against over-fitting, we also added L2 regularization on the weights.

\subsection{Training}
For training the network, as is quite common in training deep networks, we used Adam optimizer \cite{kingma2014adam} with $\beta_1=0.9$, $\beta_2=0.999$ (default values) with a batch size of $m=100$ with early-stopping (i.e., stopped training when accuracy on the test set started to decrease) to stop training.

\section{Experiments and Results}

In this section, we evaluate (i) the training and testing performance of CINet on the artificial dataset, (ii) how well CINet generalizes to scenarios where the number of contexts is more than the network is trained to, and (iii) how well CINet performs on a real scene dataset where there are scenes of different categories, which roughly correspond to different contexts.

\subsection{CINet Training and Testing Performance}

The training and testing accuracies of CINet are listed in Table \ref{tbl:accuracies}. We have experimented with different hidden memory units and layers. We also evaluated different types of inputs for the network: (i) the probabilities of contexts given objects, i.e., $\mathbf{p}_{c_i}$, (ii) the probabilities of objects given contexts, i.e., $\mathbf{p}_{o_i}$, and (iii) concatenation of $\mathbf{p}_{c_i}$ and $\mathbf{p}_{o_i}$ -- these terms were introduced in Section \ref{sect:dataset}. The number of hidden units in each layer is empirically selected as 50 (results not provided here for the sake of space). 

We see that LSTM units with 3 layers on $\mathbf{p}_{c_i}$ perform best. The better performance with $\mathbf{p}_{c_i}$ suggests that how good a context can be predicted given an object (i.e., $p(c|o)$) gives crucial information for whether to add a new context. From the table, we also see that the difference between the training and testing accuracies are rather small in cases where $\mathbf{p}_{c_i}$  and concatenation of $\mathbf{p}_{c_i}$ and $\mathbf{p}_{o_i}$ are used, suggesting that the network does not exhibit over-fitting in these cases. The fact that there is larger difference in case $\mathbf{p}_{o_i}$ implies that $\mathbf{p}_{o_i}$ is more complex and more difficult for the network to learn from.

\subsection{Applying CINet to incremental context modeling}

In this part, we evaluate CINet (with 3-layer LSTM) on the artificial test data (constructed as described in Section \ref{sect:dataset}) and on a real dataset, the SUN-RGBD dataset \cite{song2015sun}.

\subsubsection{Artificially Generated Dataset}

\begin{figure*}
\centerline{
\subfigure[Ground truth=5]{
	\includegraphics[width=0.3\textwidth]{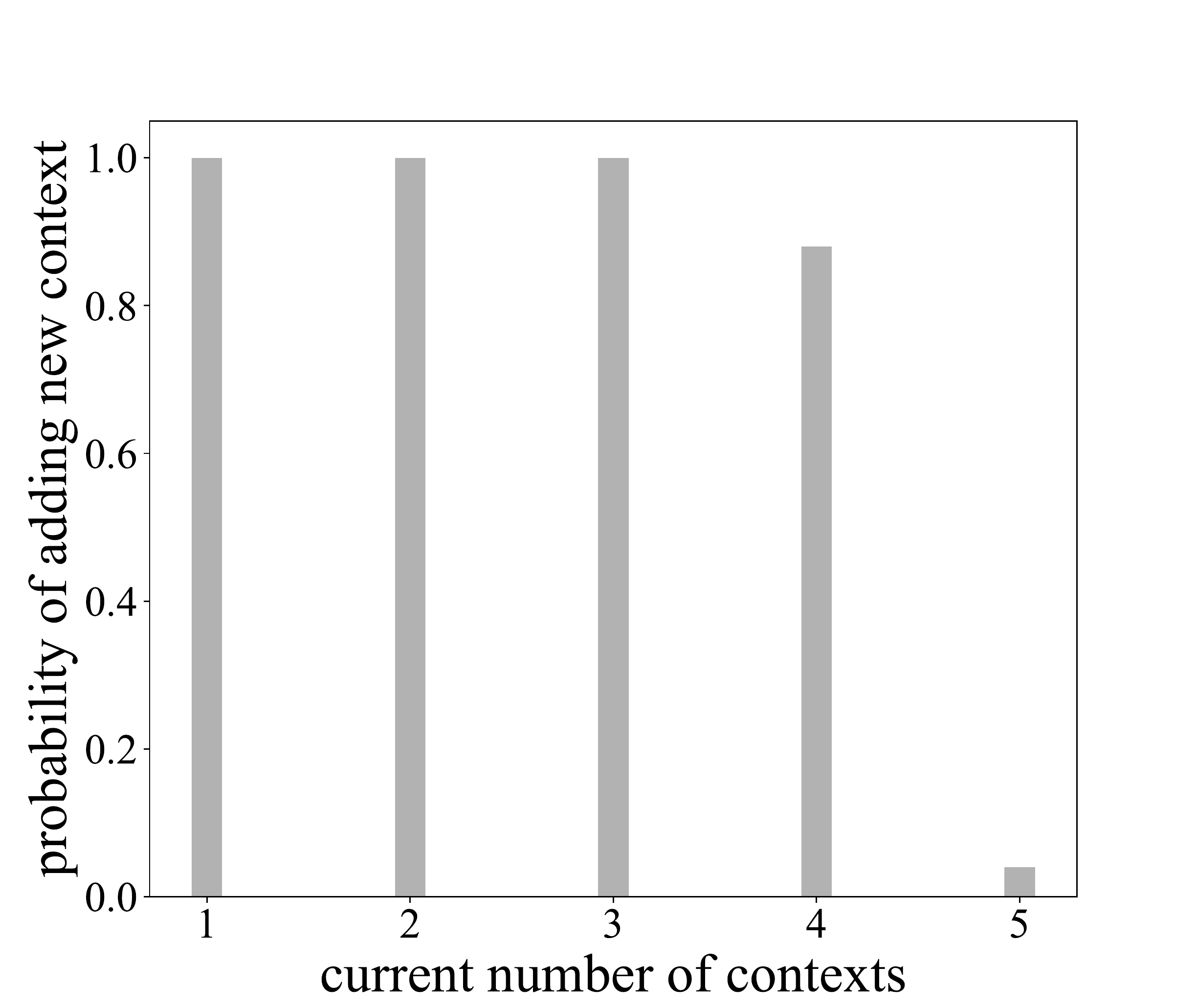}
	}
\subfigure[Ground truth=7]{
	\includegraphics[width=0.3\textwidth]{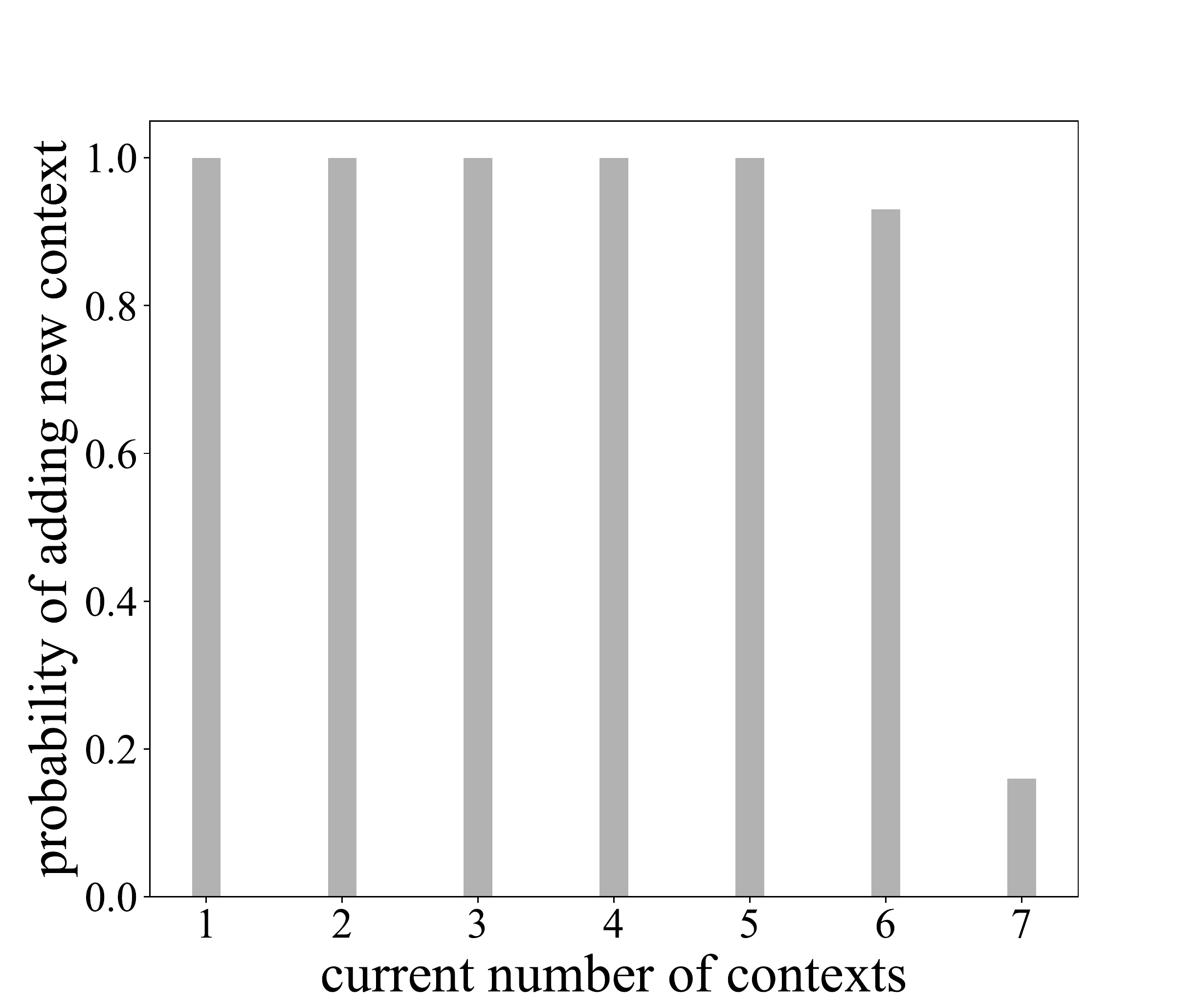}
	}    
\subfigure[Ground truth=10]{
	\includegraphics[width=0.3\textwidth]{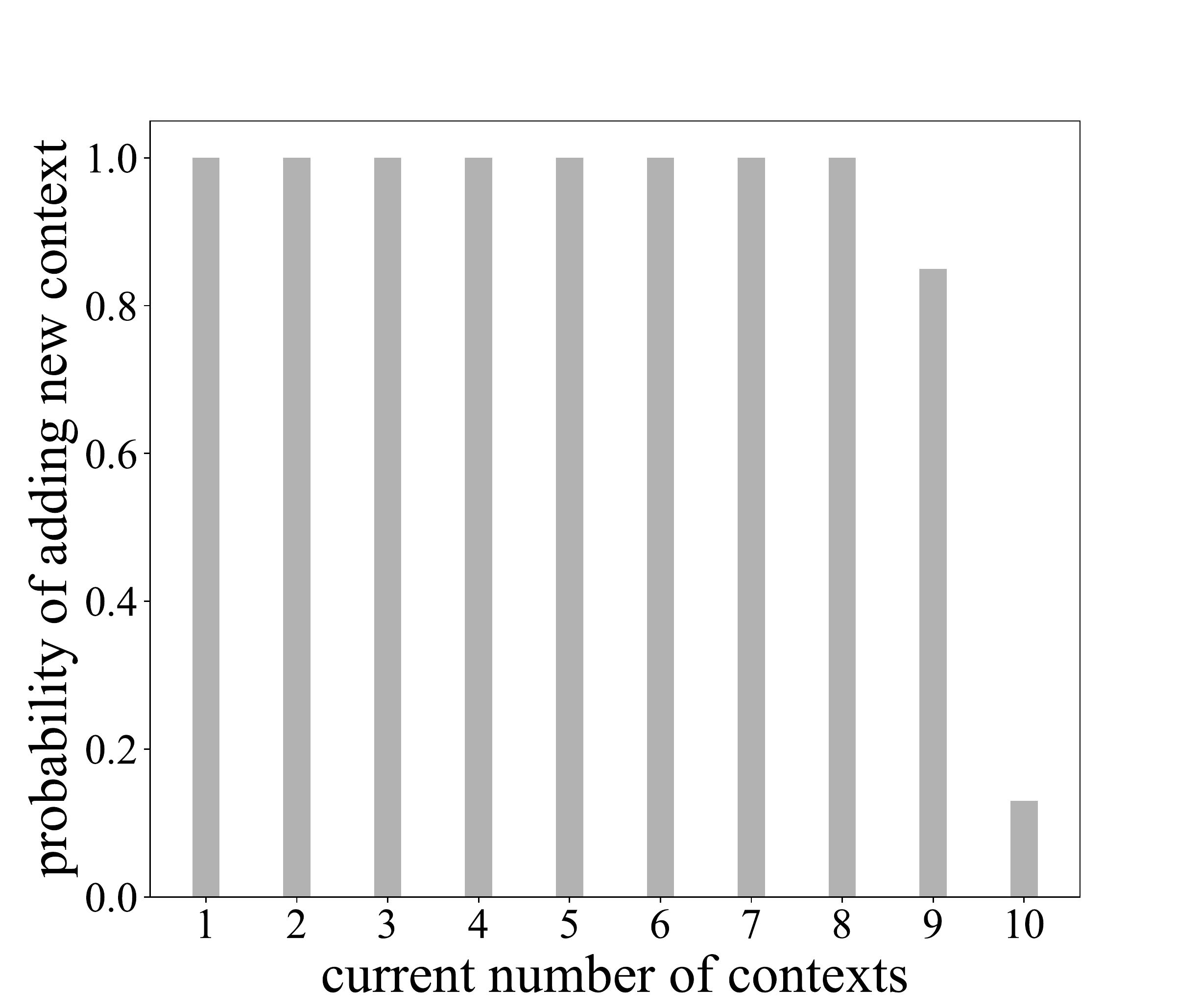}
	}
}
\centerline{
\subfigure[Ground truth=15]{
	\includegraphics[width=0.3\textwidth]{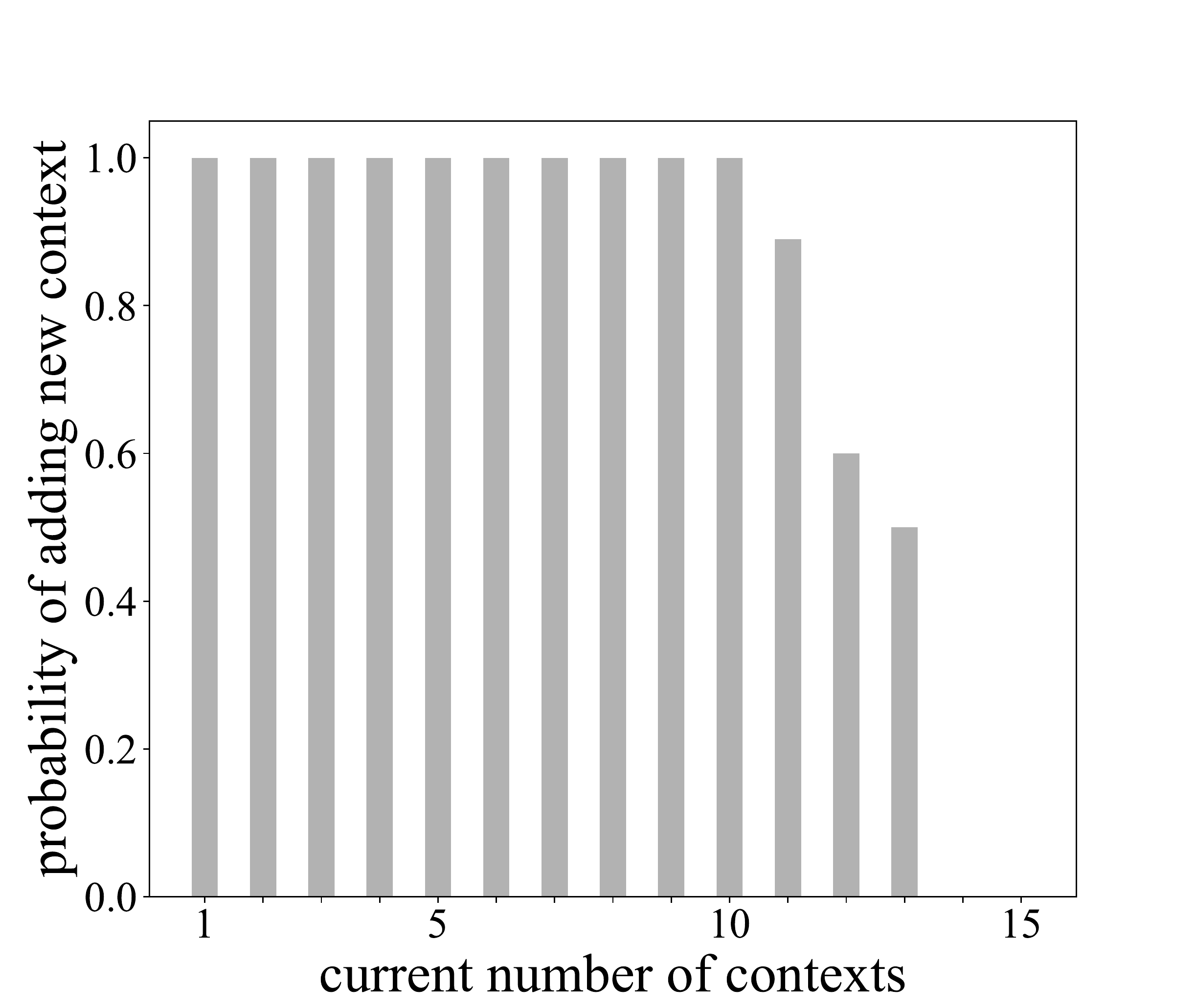}
	}        
\subfigure[Ground truth=20]{
	\includegraphics[width=0.3\textwidth]{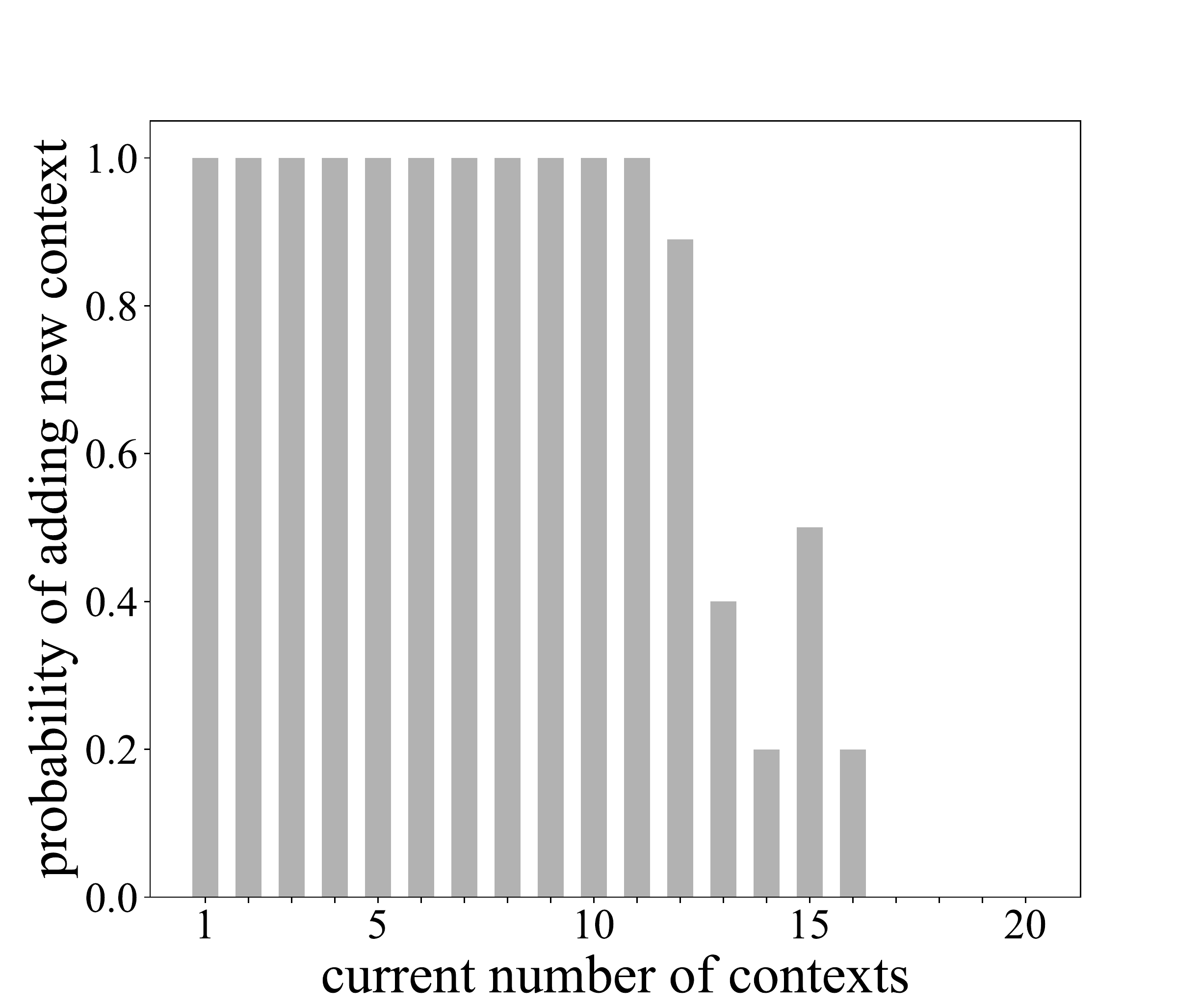}
	}        
}
\caption{Probability of incrementing contexts for various states of an LDA model on the artificial data. Ground truth is respectively (a) 5, (b) 7, (c) 10, (d) 15 and (e) 20. Note that the network was trained for LDA models up to 10 contexts.\label{fig:p_vs_topics_artificial}}
\end{figure*}

Figure \ref{fig:p_vs_topics_artificial} plots the probability of network to increment the number of contexts for various current number of contexts ($k_0$) and ground truth contexts ($k=5,7,10,15,20$ -- selected arbitrarily). We see that the network predicts to increment $k_0$ when $k_0 < k$ with very high probability (on average 0.98 for contexts $k=5,7$ and $10$; and 0.84 for $k=15$ and $20$). Moreover, as expected, when $k_0$ is close or equal to $k$, the network decreases its prediction probability. From this, we conclude that the network can nicely determine when to increase the number of contexts and when to stop.

In addition, in Figures \ref{fig:p_vs_topics_artificial}(d-e), we see the predictions when $k=15$ and $k=20$. Note that our network has been trained for artificially created data with $k$ up to $10$. The results in Figures \ref{fig:p_vs_topics_artificial}(d-e) suggest that our network generalizes well to number of contexts that is larger than what it has been trained for, although settling for a slightly lower number of contexts. 

This nice generalization behavior can be explained by the fact that the data for $k<10$ and $k>10$ follow similar distributions, and the network, when, given the probabilities between contexts and objects, is not affected by the number of contexts thanks to its weight-sharing mechanism over time steps.

\begin{figure}
\centerline{
\includegraphics[width=0.49\textwidth]{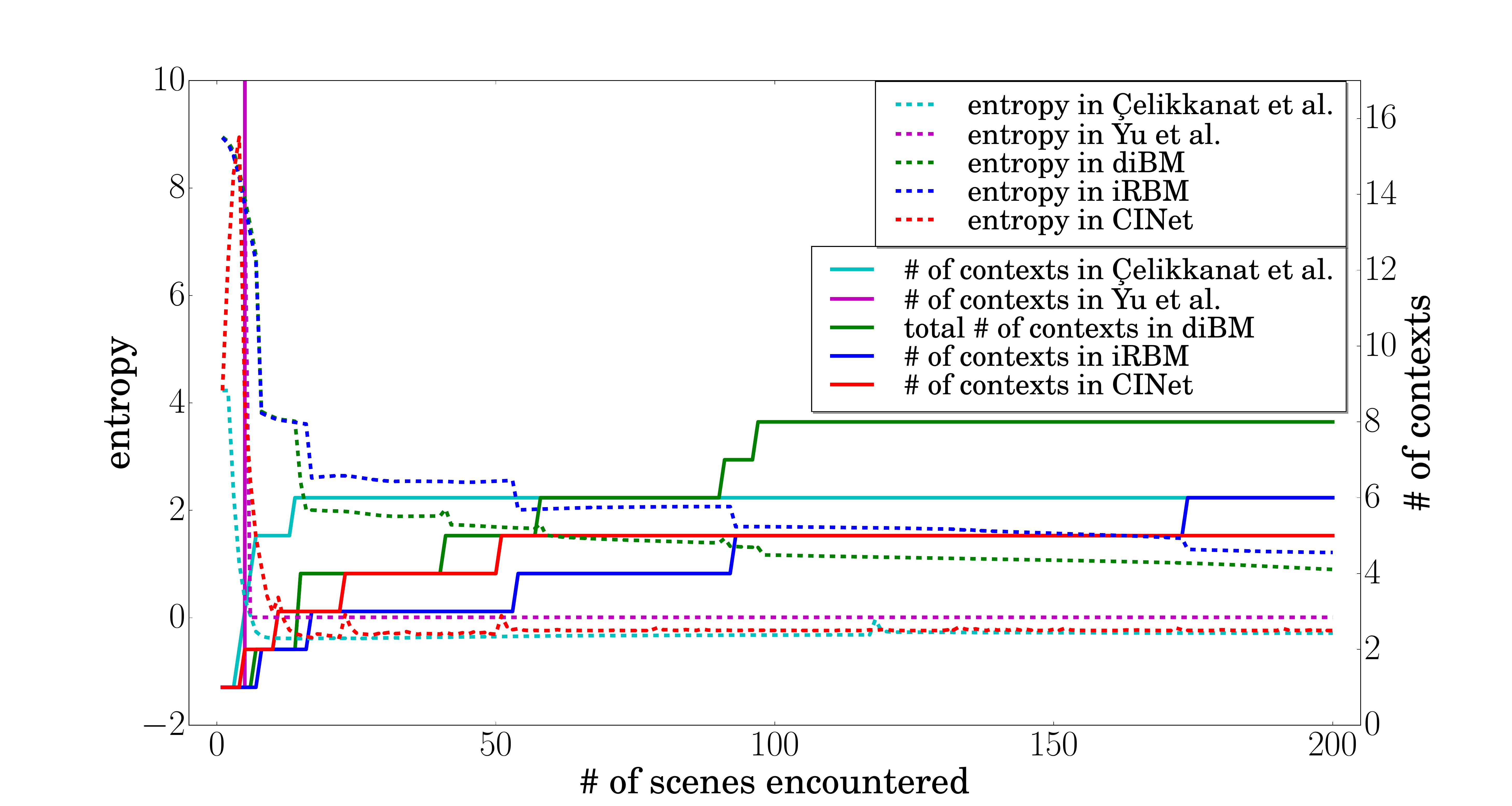}
}
\caption{How the entropy of the system changes on the \textbf{artificial} dataset with respect to change in number of contexts. The graph is for the subset of the data with 5 contexts (arbitrarily selected from the dataset). Note that the model of \cite{CelikkanatContext2014} converges to wrong number of contexts. iRBM and diBM are from \cite{DoganEtAl2018}. \label{fig:entropy_artificial}}
\end{figure}

We also analyzed how the system behaves when a new context is added, as shown in Figure \ref{fig:entropy_artificial}. Here, comparing with the result of Celikkanat et al. \cite{CelikkanatContext2014} and Dogan et al. \cite{DoganEtAl2018} (iRBM and diBM models), we see that our system converges to the correct number of contexts and yields the same level of entropy for the system, where entropy is defined as follows (as in \cite{CelikkanatContext2014}): 
\begin{equation}
\hat{H} = \rho H(o | c) + (1 - \rho )  H(c | s),
\end{equation}
{\noindent}where random variables $o$, $c$ and $s$ denote objects, contexts and scenes respectively; $H(\cdot|\cdot)$ denotes conditional entropy; and $\rho$ is a constant (selected as 0.9) controlling the importance of the two terms. The first term measures the system's confidence in observing certain objects given a context, and the second one promotes context confidence given a scene.

\subsubsection{Real Dataset}

\begin{figure}
\centerline{
	\includegraphics[width=0.3\columnwidth]{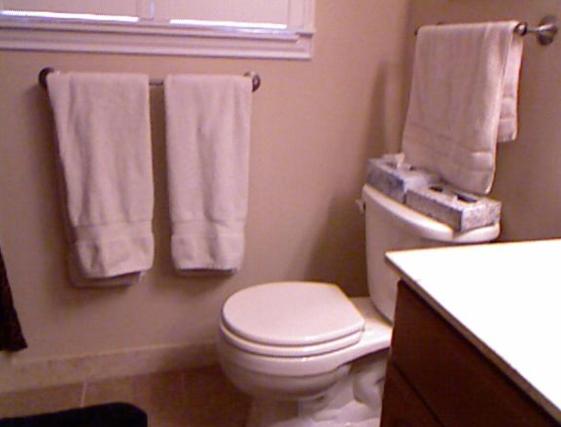}
	\includegraphics[width=0.3\columnwidth]{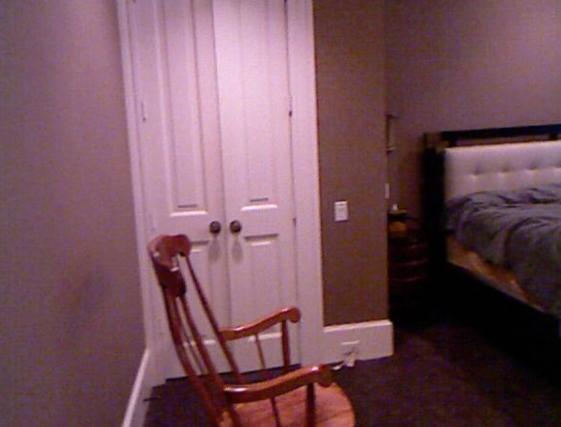}
	\includegraphics[width=0.3\columnwidth]{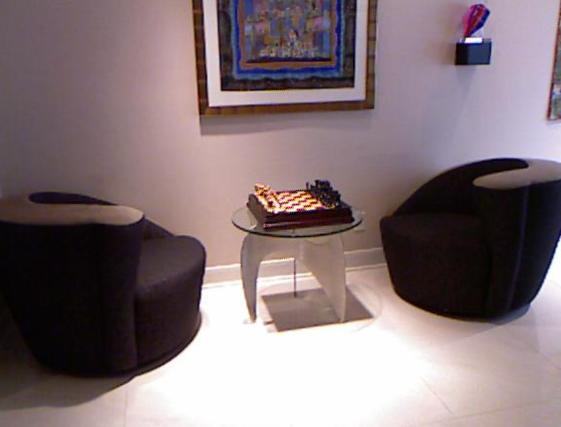}
}
\caption{A few samples from the SUN-RGBD dataset \cite{song2015sun}. \label{fig:real_dataset}}
\end{figure}

In this experiment, on a real dataset, we evaluate the model trained on the artificial dataset. We selected 878 scenes with 8 main and 25 sub-categories\footnote{A main category is, e.g., office, and its sub-category is home-office.} (the number of sub-categories gives us a baseline for the number of contexts) and 1000 objects from the SUN-RDBD dataset \cite{song2015sun} -- see Figure \ref{fig:real_dataset} for some samples. These datasets are generally used for scene segmentation and classification. We chose this dataset since it depicts a prominent challenge for robots continually interacting with different environments: The robot needs to learn the different types (contexts) of environments but does not know beforehand what and how many they are. The dataset has the objects labeled with bounding boxes on them. Although objects being labeled simplifies the task for us, object detection can easily be performed with great performances using deep learning nowadays and we leave this out of focus of the current paper.

\begin{figure}
\centerline{
\includegraphics[width=0.3\textwidth]{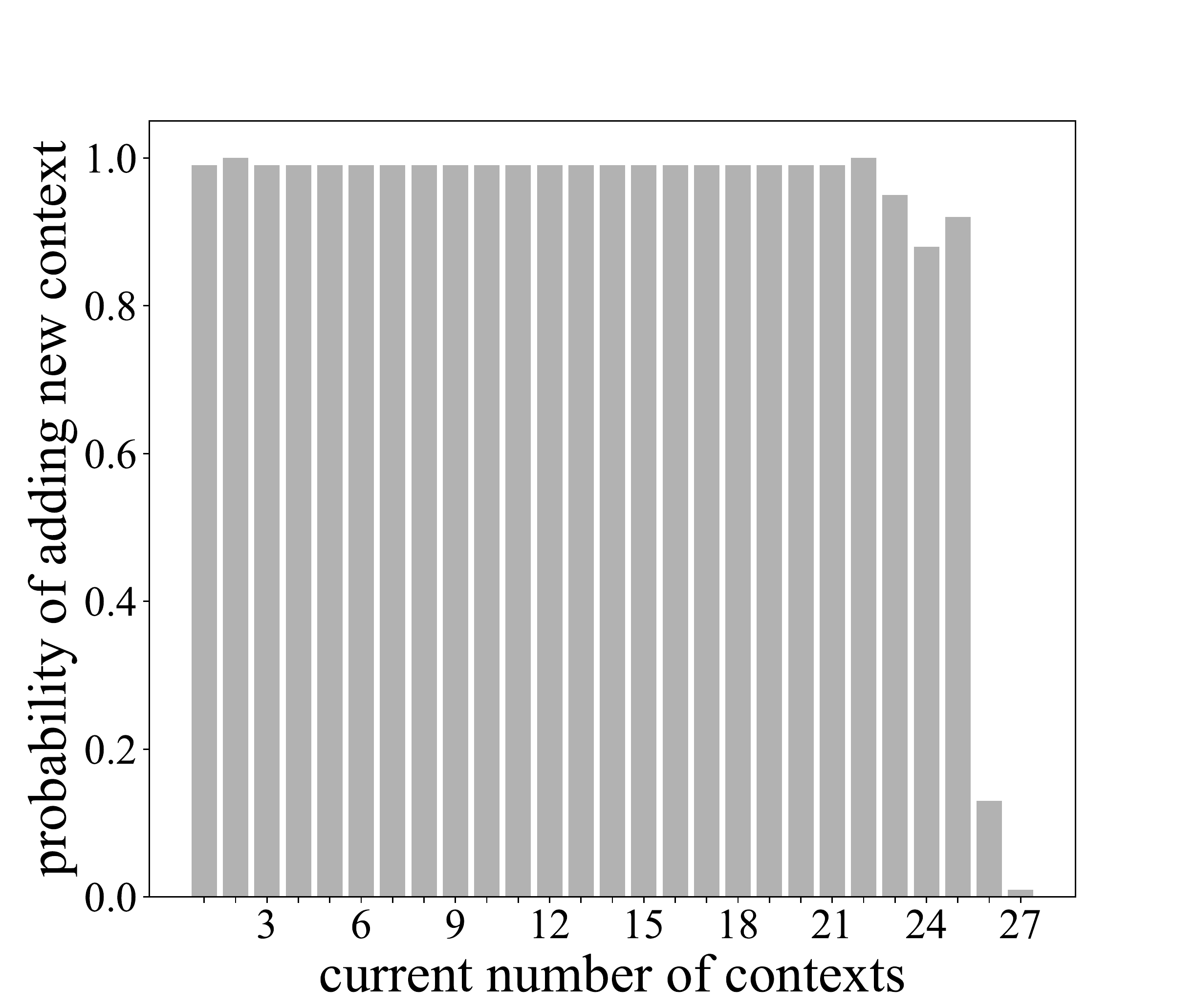}
}
\caption{Probability of incrementing contexts for various states of the LDA model on the real data. \label{fig:p_vs_topics_real}}
\end{figure}

Figure \ref{fig:p_vs_topics_real} plots how confident the network is in adding a new context for various number of contexts. We see that, when it is close to 25 (the number of sub-categories in the dataset), the probability decreases significantly, signaling the robot to not add a new context. 

\begin{figure}
\centerline{
\includegraphics[width=0.49\textwidth]{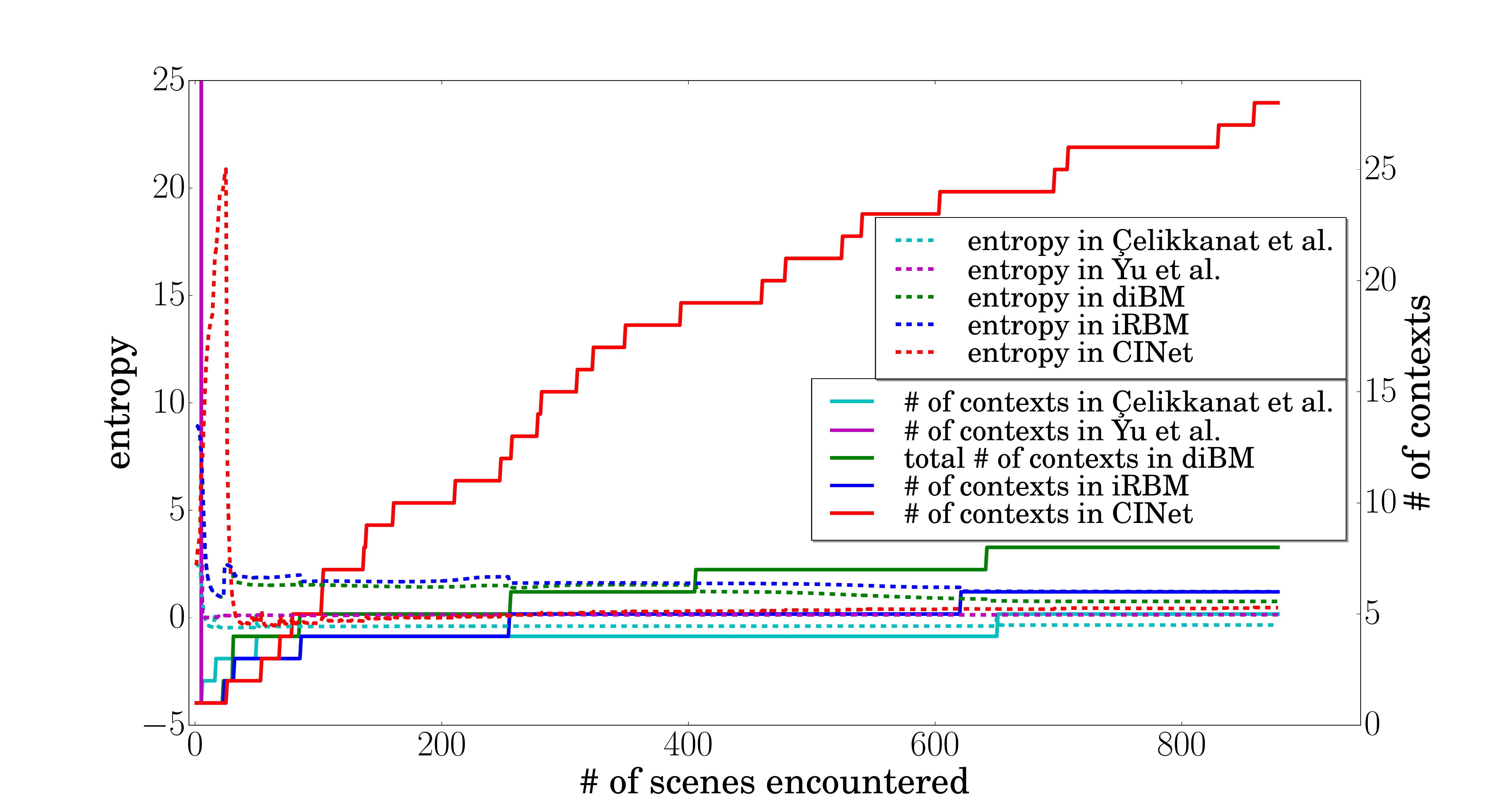}
}
\caption{How the entropy of the system changes on the \textbf{real} dataset with respect to change in number of contexts. There are 25 sub-categories (giving as a baseline for contexts). Note that the rule-based methods \cite{CelikkanatContext2014,DoganEtAl2018} diverged significantly. iRBM and diBM are from \cite{DoganEtAl2018}. \label{fig:entropy_real}}
\end{figure}

When we look at how the entropy of the system changes during encountering new scenes plotted in Figure \ref{fig:entropy_real}, we see that our system gets closer to the number of sub-categories compared to the rule-based methods \cite{CelikkanatContext2014,DoganEtAl2018}. Since the real dataset can be noisy due to labeling and the scenes might have more contexts than labeled, it is hard to expect exactly 25 contexts on the real dataset.

These results are very important since CINet was \textit{not} trained on the real dataset. These results suggest that, although the network was trained on an artificially generated dataset, since the distribution of the artificial data matches that of the real data and since it learned well to capture the distribution between contexts and objects and when a new context is needed, it generalizes well to problems with similar distributions.

\section{Conclusion}

We have proposed a learning based approach to incremental model building on robots continually interacting with new types of environments (contexts). To the best of our knowledge, this is the first work formulating this as a learning problem. We have used Latent Dirichlet Allocation to generate data with known contexts and trained Recurrent Neural Networks to estimate when to add a new context. We evaluated our system on artificial and real datasets, showing that the network performance well on test data (98\% accuracy) with good generalization performance on the real dataset.\\
Our work can be extended by, e.g., defining context on more detailed scene models (e.g., with spatial, temporal or categorical relations), adapting the model for different types of contexts (e.g., social, temporal etc.), and formulating construction of a latent hierarchy as a learning problem as well.
\vspace{-0.8em}
\section*{Acknowledgment}
This work was supported by the Scientific and Technological Research Council of Turkey (T\"UB\.{I}TAK) through project called ``Context in Robots'' (project no 215E133). We gratefully acknowledge the support of NVIDIA Corporation with the donation of the Tesla K40 GPU used for this research.

\bibliographystyle{IEEEtran}
\bibliography{references}

\end{document}